\documentclass[10pt,twocolumn,letterpaper]{article}

\usepackage[pagenumbers]{wacv} 

\usepackage{graphicx}
\usepackage{amsmath}
\usepackage{amssymb}
\usepackage{booktabs}
\usepackage{comment}
\usepackage{mathtools}
\usepackage[accsupp]{axessibility}
\usepackage[pagebackref,breaklinks,colorlinks]{hyperref}

\usepackage[capitalize]{cleveref}
\crefname{section}{Sec.}{Secs.}
\Crefname{section}{Section}{Sections}
\Crefname{table}{Table}{Tables}
\crefname{table}{Tab.}{Tabs.}

\def\wacvPaperID{1374} 
\def\confName{WACV}
\def\confYear{2024}
\newcommand{\vect}[1]{\boldsymbol{\mathbf{#1}}}

\begin{document}

\title{A Neural Height-Map Approach
for the Binocular Photometric Stereo Problem}

\author{Fotios Logothetis\\
\small Cambridge Research Laboratory,Toshiba Europe Ltd.\\
\small Cambridge, UK \\
{\tt\small fotios.logothetis@toshiba.eu}
\and
Ignas Budvytis \\
\small University of Cambridge\\
\small Cambridge, UK \\
{\tt\small ib255@cam.ac.uk }
\and
Roberto Cipolla \\
\small University of Cambridge \\
\small Cambridge, UK\\
{\tt\small rc10001@cam.ac.uk}
}
\maketitle

\begin{abstract}
   In this work we propose a novel, highly practical, binocular photometric stereo (PS) framework, which has same acquisition speed as single view PS, however significantly improves the quality of the estimated geometry. 

As in recent neural multi-view shape estimation frameworks such as NeRF~\cite{mildenhall2020nerf}, SIREN~\cite{sitzmann2019siren} and inverse graphics approaches to multi-view photometric stereo (e.g. PS-NeRF~\cite{psnerf}) we formulate shape estimation task as learning of a differentiable surface and texture representation by minimising surface normal discrepancy for normals estimated from multiple varying light images for two views as well as discrepancy between rendered surface intensity and observed images. Our method differs from typical multi-view shape estimation approaches in two key ways. First, our surface is represented not as a volume but as a neural heightmap where heights of points on a surface are computed by a deep neural network. 
Second, instead of predicting an average intensity as PS-NeRF or introducing lambertian material assumptions as Guo et al.~\cite{guo2022edgepreserving}, we use a learnt BRDF and perform near-field per point intensity rendering. 

Our method achieves the state-of-the-art performance on the DiLiGenT-MV dataset adapted to binocular stereo setup as well as a new binocular photometric stereo dataset - LUCES-ST.
\end{abstract}

\section{Introduction}
\label{sec:intro}

\begin{figure}[t]
    \centering
  \includegraphics[width=\columnwidth]{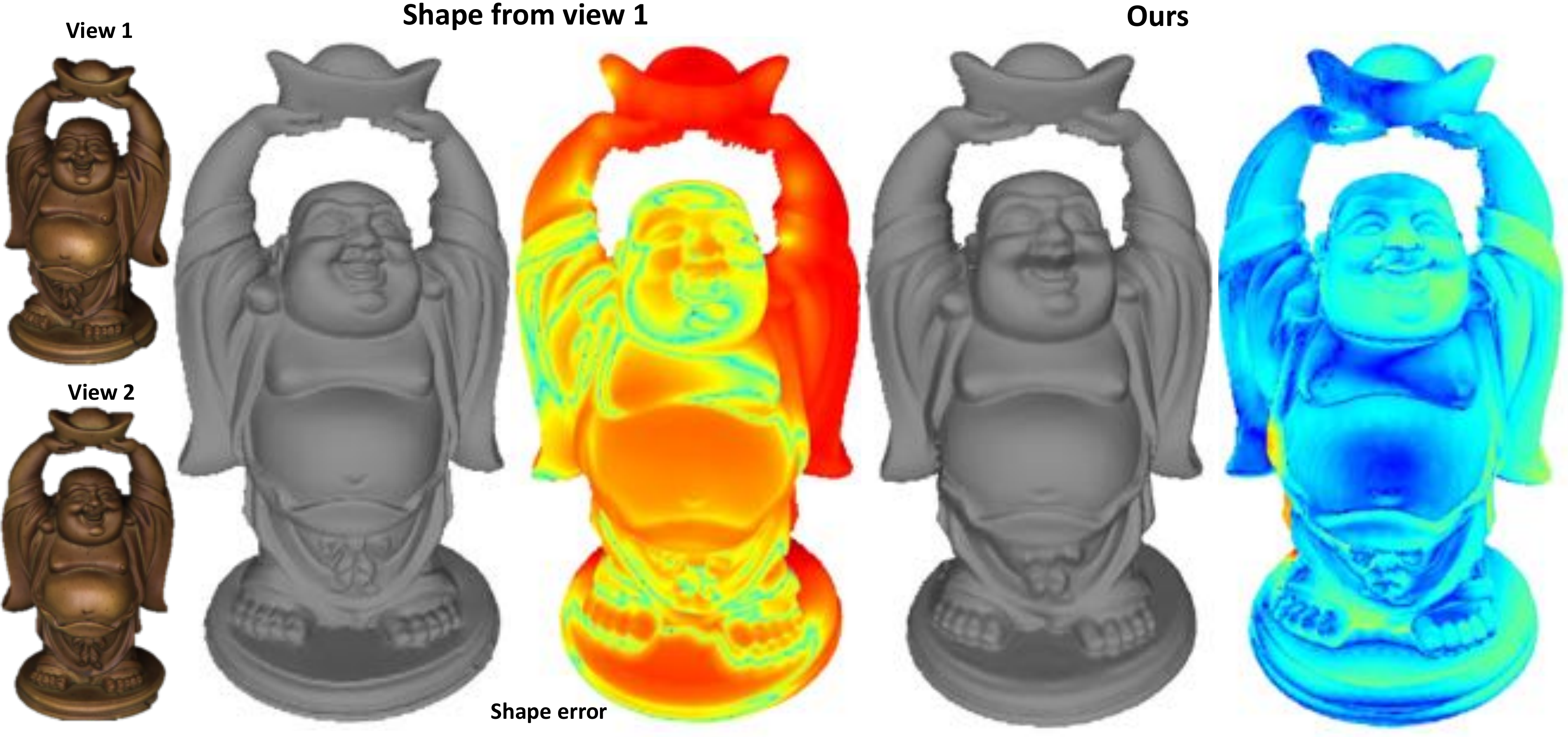}
    \caption{This figure illustrates the limitations of the use of the single view photometric stereo for shape estimation. The left pair of images show a plausible shape estimated using single view PS method of~\cite{Logothetis22} which respects estimated normals well (zoom in for a better view), however its actual shape error is large due to bending (right side of the statue) caused by shape discontinuities or small systematic errors in estimated normals. In contrast, our Binocular Photometric Stereo method while having the same capture time and little hardware costs obtains a significantly improved shape. Error maps are visualised by using jet color scheme on per-pixel shape error. All pixels with an error above 1.5mm are assigned with dark red color.} 
    \label{fig:into}
\end{figure}

Single view Photometric Stereo is a long standing problem in Computer Vision. Recent methods~\cite{guo2022edgepreserving,winpxnet,Logothetis22} have achieved impressive normal estimation accuracy on both real and synthetic~\cite{ikehata2018cnn,Logothetis22} datasets. However, the progress in the quality and practical usefulness of the estimated shape (e.g., by using the numerical integration of~\cite{queau2015edge}) has been much less convincing, due to the heavily ill-posed nature of the global shape estimation problem (see Figure~\ref{fig:into}). 

One way to improve the quality of the global shape extracted from Photometric Stereo images is to leverage multiple views. A classical approach to multi-view photometric stereo~\cite{LiZWSDT20} involves obtaining initial sparse point-cloud via structure from motion~\cite{schoenberger2016sfm}. Depths of these points are then propagated along the iso-depth contours in each view. Not having any learnable component this approach is fragile to errors in computation of the iso-depth contours. Recently proposed PS-NeRF~\cite{psnerf} reformulated the multi-view photometric stereo problem as an inverse graphics problem by learning a neural textured volume to minimise discrepancy between estimated surface normals and predicted photometric stereo normals as well as predicted intensities and observed intensities. 
It unsurprisingly achieves the state-of-the-art on the DiLiGenT-MV~\cite{LiZWSDT20} multi-view photometric stereo benchmark. 

While multi-view photometric stereo can achieve low reconstruction errors (few tenths of milimeters), for some applications such as robotic interaction, and conveyor belt scanning, it is not feasible due to long capture times and precise camera pose calibration (especially when a single camera is used) required. Hence in this work we consider the binocular photometric stereo setup where multiply lit images are obtained for a pair of cameras. Note that Binocular Photometric Stereo has been introduced in \cite{KongXT06} and subsequently developed in \cite{DuGS11,WangWMHCS13} under different assumptions and for different applications. However the aforementioned methods do not model materials with complex reflectances, e.g., highly specular materials such as metal or porcelain.

To address this limitation, we adapt the recently popular neural rendering approaches (e.g. NeRF~\cite{mildenhall2020nerf}, SIREN~\cite{sitzmann2019siren}, PS-NeRF~\cite{psnerf}) to the binocular photometric stereo setup. Note while it is a popular belief 
that NeRF-like approaches do not work well in sparse setup we show that two views are enough to compute accurate shape (see Tables~\ref{tab:Tab_eval_diligent} and~\ref{tab:tab_eval_stereoluces}) if care is taken in modelling of the neural representation of shape and losses used. In particular, instead of a neural density~\cite{mildenhall2020nerf} or signed distance field~\cite{sitzmann2019siren} we leverage a neural heightmap where heights of points on a surface are computed by a deep neural network.
In comparison to volume-based shape representations, this allows for better conditioned and efficient surface optimisation procedure. Moreover, instead of predicting an average intensity as PS-NeRF\footnote{Note, PS-NeRF~\cite{psnerf} runs in two stages. First stage uses average images and renders intensities per light image only in the second stage during which the shape is not updated.} or introducing lambertian material assumptions as Guo et al.~\cite{guo2022edgepreserving},  we use a learnt BRDF and perform near-field per point intensity rendering. 

In more detail, our method works by combining three steps of: (1) estimating per-view based shape by using per view estimated photometric stereo normals~\cite{Logothetis22} and (2) using it to initialize neural heightmap network guided by estimated pixel-wise normals and depth (initialising the albedo value to a constant) and (3) fitting the initialised neural heightmap to image intensity and estimated normal maps. 

Our method achieves the state-of-the-art performance on the DiLiGenT-MV~\cite{LiZWSDT20} dataset adapted to binocular stereo setup. It is also evaluated on a new binocular photometric stereo dataset, LUCES-Stereo, consisting of 7 objects from original LUCES~\cite{Mecca21} dataset captured in the binocular photometric stereo setup. See Sections~\ref{sec:LUCES} and~\ref{sec:experiments} for more details.
Our contributions include: \begin{itemize}
     \item A neural height-map approach to the Binocular Photometric Stereo problem which is robust to highly complex materials
     \item A Binocular Photometric Stereo dataset - LUCES-ST.
 \end{itemize}

\newpage
\section{Related Work}
\label{sec:relatedworks}

There is an extensive literature leveraging photometric cues for single and multi-view based 3D reconstruction. 
Here  we categorise as follows.

\noindent
\textbf{Single view photometric stereo.} 
Recently deep PS has been very successful on solving the single view far-field PS problem from CNN-PS   \cite{ikehata2018cnn} to PX-Net~\cite{logothetis2021pxnet} which is also extended for the near-field setting \cite{Logothetis22}. Other works  like \cite{YakunLearning2023} incorporated  material reflectance priors for single view normal prediction or used specific BRDFs \cite{GoldmanCHS10,EstebanVC08,mecca2016single,logothetis2016near,logothetis2017semi}, including Lambertian or Ward reflection models. Other recent approaches have also tackled a more uncalibrated setting like \cite{chen2019SDPS_Net,li2022selfps,li2022neural,yang2022snerf}. Finally, \cite{guo2022edgepreserving} introduced the idea of a infinitely differentiable surface (SIREN \cite{sitzmann2019siren}) with Lambertian rendering to directly optimise a neural surface from intensities. Our method is similar to \cite{guo2022edgepreserving} but extended in a stereo setting and with a non-Lambertian rendering.


\noindent
\textbf{Sensor enhanced photometric stereo.}
Some works have also utilised various 3D scanning techniques such as laser scanner and structured light~\cite{LevoyPCRKPGADGSF00,RusinkiewiczHL02,ZhangSCS04} allowing to fit reflectance functions at each surface point. While it may be possible to combine photometric stereo with structured light scans \cite{NehabRDR05,AliagaX08}, accurately merging RAW data from different type of scans is a challenging task and can limit the resolution of the reconstruction



\begin{figure*}[ht]
    
    \begin{center}\includegraphics[width=0.9\textwidth]{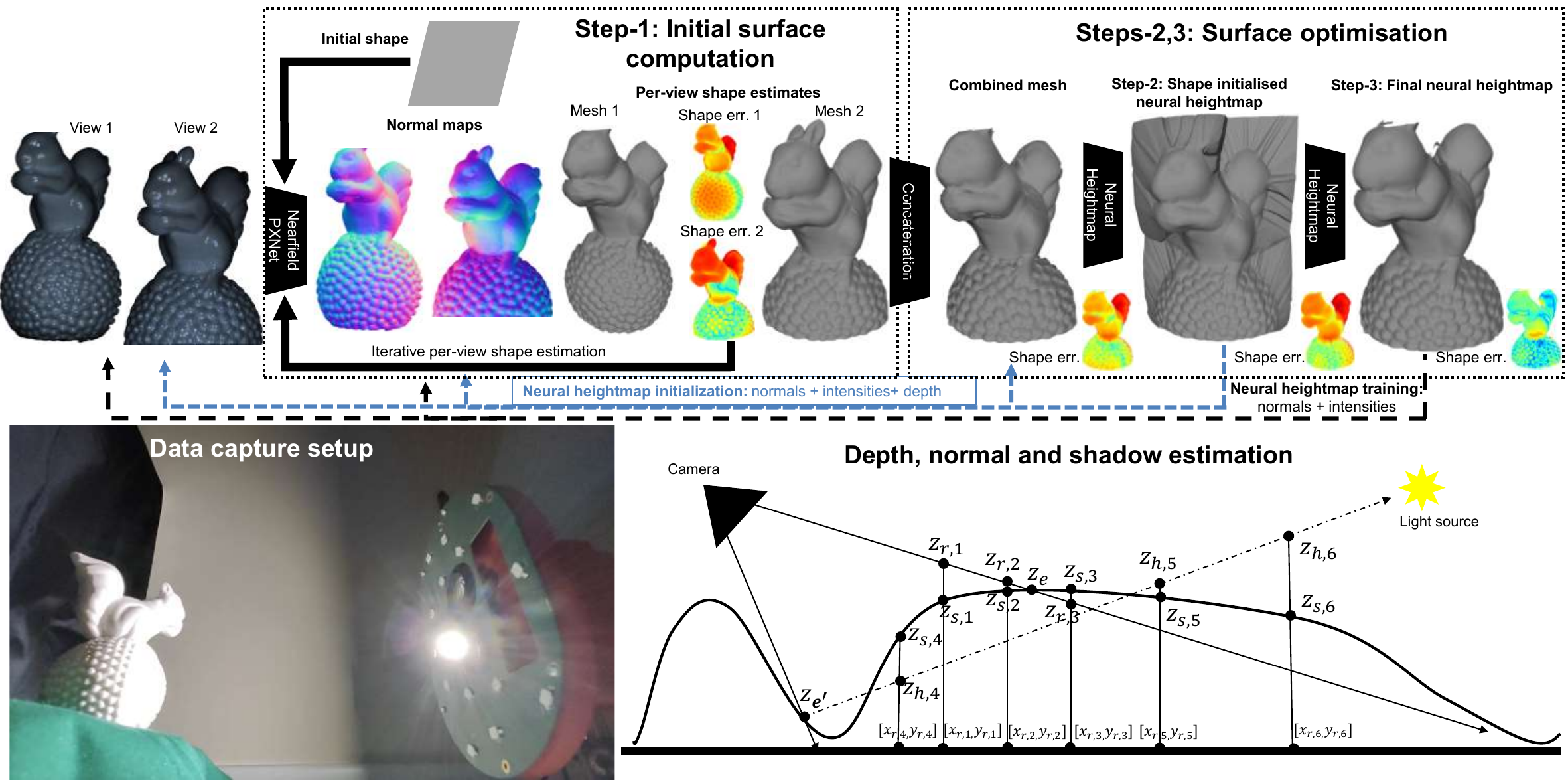}\end{center}\vspace{-0.4cm}
    \caption{Graphic illustration of the proposed approach. The bottom left shows our binocular photometric stereo data capture setup. The top figure illustrates three key steps of our method: (1) joint normal and shape estimation for each view using~\cite{Logothetis22}, (2) initialisation the neural heightmap based on SIREN~\cite{sitzmann2019siren} architecture using the shape and normals estimated in step 1 and (3) the main training step of the neural heightmap. The bottom right part of this figure illustrates the ray sampling procedure used to compute normal, depth and shadow estimates from the heightmap. Note that sample points 1-3 and 4-6 correspond to 2 different surface points $z_e$ and $z_{e^{'}}$. }
    \label{fig:method}
\end{figure*}



\noindent
\textbf{Binocular photometric stereo.} Specific binocular Photometric Stereo has been introduced in \cite{KongXT06} and subsequently developed in \cite{DuGS11,WangWMHCS13} under different assumptions and for different applications. The limitations in these cases are the lack of generality in terms of material reflectance which makes these methods not being very effective with specular outliers.

\noindent
\textbf{Multi-view photometric stereo.} \cite{Logothetis19} proposed a multi-view Photometric Stereo which retrieved the volume with the sign distance function based parameterisation \cite{NiessnerZIS13,ZollhoferDIWSTN15}. Such approach relied on structure-from-motion initialisation and the photometric refinement used diffuse image irradiance equations.

Similarly, \cite{LiZWSDT20} introduced a method for capturing both 3D shape and reflectance with a multi-view photometric stereo setup. The idea is to collect photometric stereo images multiple viewpoints and combine it with structure-from-motion to obtain a precise reconstruction of the complete 3D shape. The spatially varying isotropic bidirectional reflectance distribution function (BRDF) is captured by simultaneously inferring a set of basis BRDFs and their mixing weights at each surface point.

Recently, neural surface approaches have become very popular from the introduction of NeRF \cite{mildenhall2020nerf}. This has been extended to neural SDF approaches like   \cite{Kaya2021,kaya2022uncertainty,kaya2023multi} and very recently to more structured rendering approaches like Ref-NeRF \cite{refnerf} and PS-NeRF \cite{psnerf}.

\newpage
\section{Method}
\label{sec:method}

Our binocular photometric method consists of three key steps. First step involves joint normal and shape estimation for each view indepentendly and is described in Section~\ref{sec:normalest}. Second step uses the estimated shape to initialise a neural heightmap described in Section~\ref{sec:heightmap}. Finally, the initialised neural heightmap is trained, using losses described in Section~\ref{sec:losses} to explain observed photometric stereo image intensities and estimated normals as explained in Section~\ref{sec:training}.

\noindent
\subsection{Per-view shape estimation} \label{sec:normalest}

We start by computing per view normal maps using the state-of-the-art PS normal estimation network - PX-Net \cite{Logothetis22}. This method offers a general near-field network that obtains high quality normal maps for the calibrated, near (and far) field PS setting as well as some reasonable surface estimate. 
Qualitative examples of the shape obtained using~\cite{Logothetis22} for camera 1 are shown in Figure~\ref{fig:stereoresults} (see column \textit{Logothetis et al.}). We use the estimated per view shape to initialise neural heightmap as described in Section~\ref{sec:training}. Note, as shown in Figures~\ref{fig:method} and~\ref{fig:stereoresults}, high quality local shape is obtained whilst suffering from global bending due to the ill-posed nature of shape estimation under discontinuities or systematic error in estimated normals. Note the step of per view shape from normal estimation is crucial to speeding up the recovery and constraining of the neural surface as purely relying on PS image intensities from sparse viewpoints is likely to take a significant training time and obtain suboptimal surface as indicated by some results discussed in Section~\ref{sec:experiments}.



\subsection{Neural heightmap}
\label{sec:heightmap}

\noindent
\textbf{Surface parameterisation.} We start by assuming that the surface can be expressed as a continuous height map $z_s=F(x_s,y_s)$ in some word coordinate system (we use the subscript $s$ to denote surface coordinates). For the case of stereo cameras, this coordinate system is chosen as the average between the 2 camera system (i.e. the `rectified' stereo system). We note that a roto-translation ($R_c$, $\vect{t}_c$) is required to convert between this coordinate system and the original camera coordinate system i.e.: 

\begin{equation}
[x_s,y_s,z_s]^\intercal = R_c \cdot [x_c,y_c,z_c]^\intercal + \vect{t}_c
\label{eq:coordinate}
\end{equation}

The unknown function $F$ is a deep neural network and the objective is to optimise its weights. Extending  \cite{guo2022edgepreserving} to the 2 view problem, we chose the SIREN architecture \cite{sitzmann2019siren} which is an MLP with sinusoidal activation functions and that guarantees that the surface is infinitely differentiable thus can be easily recovered from its derivatives; thus the surface normal is $\vect{n}_s \propto [ \frac{\partial F}{\partial x_s},\frac{\partial F}{\partial y_s},-1]^\intercal $ and automatic differentiation makes it a function of the network weights. We also add a scalar (grayscale) albedo $\rho= F(x_s,y_s)$ channel on the SIREN used for rendering.

\noindent
\textbf{Surface sampling}. We note that since the projection depth $z_c$ in Equation~\ref{eq:coordinate} is unknown, exact conversion between coordinate systems is impossible. This is a clear difference to the single view of \cite{guo2022edgepreserving} where one-to-one mapping between image and depth exists (this can never be the case in stereo due to left-right occlusions). To overcome this issue, a number of tentative depth samples $z_{c1}, z_{c2},  z_{c3},  \hdots$ are considered and are used to generate points $i$ along the viewing direction $\vect{v}$ as  $ \{\vect{v} z_{ci} \}$. Applying the coordinate transfer Equation~\ref{eq:coordinate} gives the coordinates of these points in the world space as $ \{[x_{ri}, y_{ri}, z_{ri}]  \}$ . This is visualised in  Figure~\ref{fig:method}. Then, the network function $F$ can be queried in the position $ \{[x_{ri}, y_{ri}] \}$ to get the surface depth estimates  $ \{ z_{si} = F[x_{ri}, y_{ri}]  \}$. Finally, in order to get the `actual' depth estimate $z_e$, the set of depth estimates is reduced with volumetric rendering, using as opacity $\alpha _i$ the inverse of the depth squared difference between  $z_{si}$ and $z_{ri}$, i.e. $\alpha _i =\exp(-f(z_{si}-z_{ri})^2)$ , with $f$ being a scaling factor  used to convert between millimeters and normalised units. The volume rendering equation then becomes:  $ z_e=  \sum_{i} \Big{(} z_{si} \alpha_i \sum_{i} \big{(} 1-\alpha_{i-1}  \big{)}   \Big{)}  $. We note that the surface normal $\vect{n}_e$ and albedo $\rho_e$ are computed with a similar volume rendering equation using the same opacity $\alpha_i$.



\begin{figure}[t]
    \includegraphics[width=0.975\columnwidth]
    {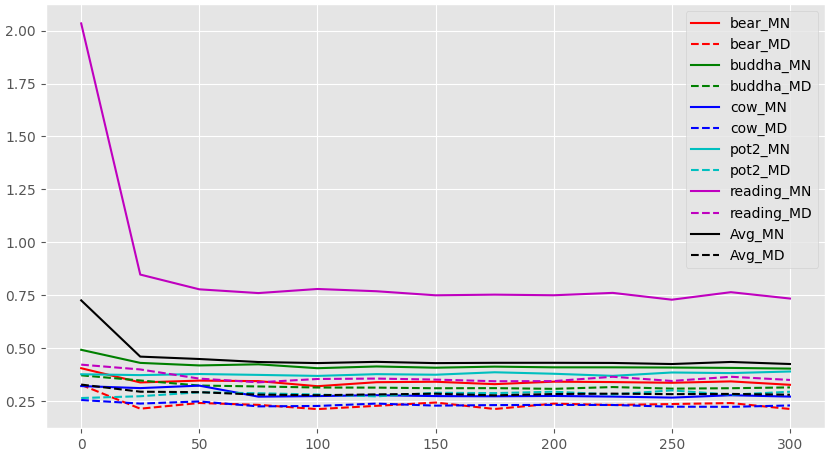}
    \vspace{-0.3cm}
    \caption{This figure plots the mean and median shape error for all DiLiGenT-MV objects. It is observed that we achieve quick convergence with minimal improvement after a few tens of epochs, where each epoch takes roughly 1 minute on an unoptimised code run on a single RTX4080 GPU.} 
    \label{fig:roccurve}
\end{figure}

\noindent
\textbf{Intensity rendering.}
To render light intensities, we first need to compute the near-field lighting vectors  $\vect{l_m}$ and light attenuation $a_m$ (for light source $m$). Following the near lighting model from \cite{Mecca2014near},  for calibrated point light sources at positions $\mathbf{s}_m$,  each surface point  $ \mathbf{p}$ gets variable lighting vectors  $\vect{l}_m=\mathbf{s}_m-\mathbf{p}$  and  attenuation factors $a_m(\mathbf{p})=\phi _m \frac{( \hat{\mathbf{L}}_m (\mathbf{X}) \cdot \hat{\mathbf{d}}_m)^{\mu _m}}{||\mathbf{l}_m (\mathbf{p})||^2}$ where $\hat{\mathbf{l}}_m=\frac{\mathbf{l}_m}{||\mathbf{l}_m||}$ is the lighting direction, $\phi _m$ is the intrinsic brightness of the light source,~$\hat{\mathbf{d}}_m$ is the principal orientation of the LED and $\mu _m$ is an angular dissipation factor.

\begin{figure*}[t]
    \begin{center}
    \includegraphics[width=0.9\textwidth]{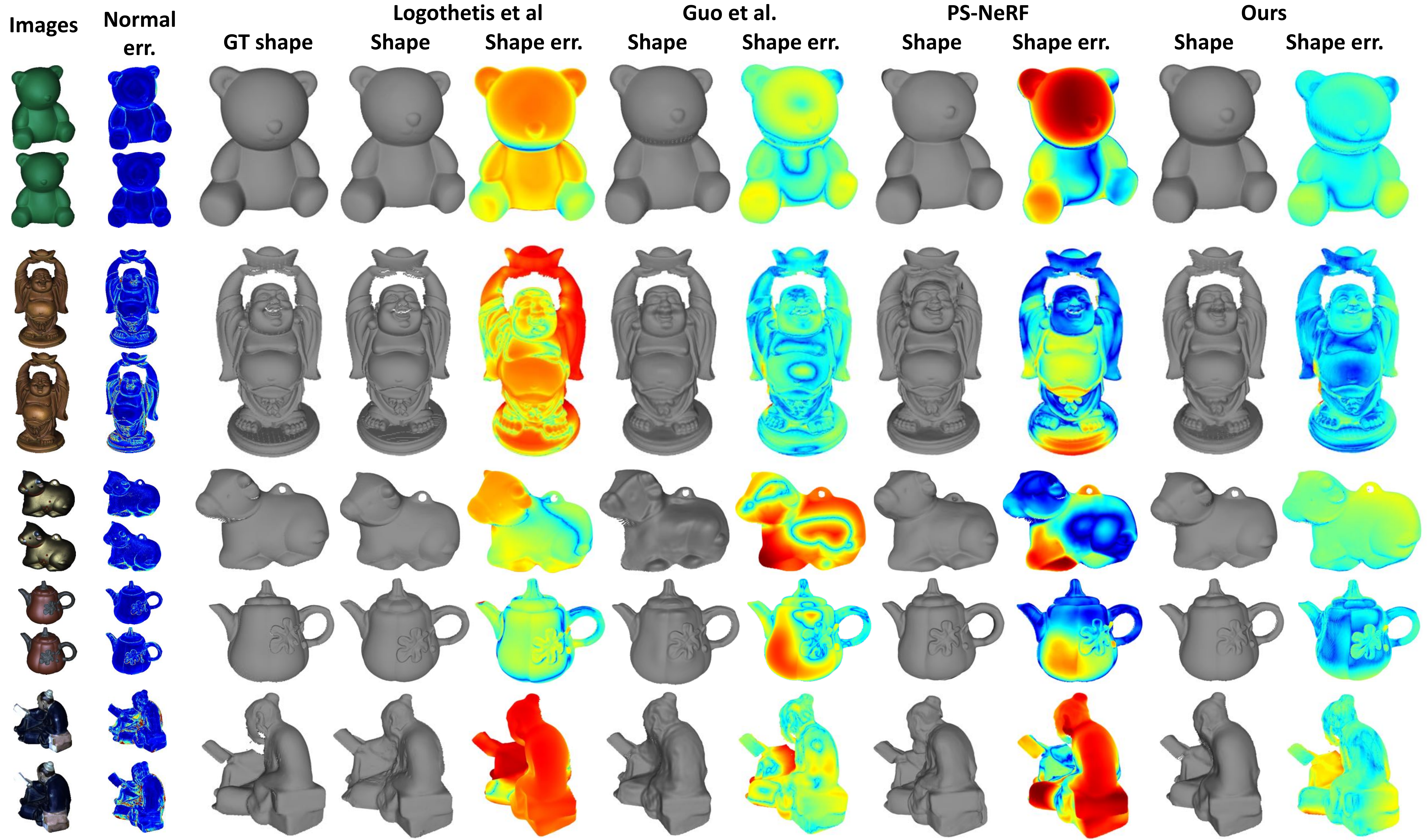}\end{center}
    \vspace{-0.4cm}
    \caption{This figure shows qualitative results on DiLiGenT-MV~\cite{LiZWSDT20} objects for our Logothetis et al~\cite{Logothetis22} (single view - camera 1), adaptation of Guo et al.~\cite{guo2022edgepreserving} to two view shape estimation, PS-NeRF~\cite{psnerf} and ours. For each of the method we visualise the predicted shape and predicted shape error map using jet color scheme (errors above 1.5mm have saturated red color). Our method outperforms all methods. Also note, that as expected, the adaptation of Guo et al.\cite{guo2022edgepreserving} fails on cow object significantly as it is a highly specular object. PS-NeRF seems to have its reconstructions divided into strongly correct (top of \textit{Pot2}) and strongly incorrect (bottom of \textit{Pot2}) regions which is likely due to the use of systematically incorrect normals. }
    \label{fig:stereoresults}
\end{figure*}

Then, the total intensity $i_m$ is computed as $i_m= s_m \cdot a_m \cdot \rho \cdot  \textbf{BRDF}(\vect{n},\vect{l_m},\vect{v})$. Here $s_m$ is a `soft' indicator variable that is 0 for shaded points and 1 otherwise (see bellow). 

\noindent
\textbf{Learned BRDF renderer.}
Our aim is to learn a single BRDF model (assuming uniform material with potentially varying albedo) following the principles described in the MERL real material database \cite{Matusik2003jul}. For that, the half vector $\vect{h}=\frac{\vect{l_m}+\vect{v}} {|\vect{l_m}+\vect{v}|} $ is first computed as well as the relatives angles between $\vect{n}, \vect{h}$ and $\vect{l}$ namely  $\theta _h, \phi _h,  \theta _d, \phi _d$ (see supplementary for more details). Moreover, it is desired to only recover isotropic materials there $\phi _h$ is ignored. In addition, real BRDFs follow the Helmholtz reciprocity constraint and so $\text{BRDF} (..., \phi _d)=\text{BRDF} (..., \phi _d +\pi)$. 

Thus, we parameterised  
\begin{equation}
\text{BRDF} (\vect{n},\vect{l_m},\vect{v}) \vcentcolon = (\vect{n}\cdot \vect{l_m}) \text{MLP} (\theta _h, \theta d, \phi _d)
\label{eq:brdf}
\end{equation}

we use 3x16 hidden layers with relu activation and exponential activation (the BRDF values must be always non-zero and should be around 1 for diffuse materials) for the output layer. We note that even though the surface point is computed as a weighted sum over the ray (as above), the renderer is only computed on a single sample. 

\noindent
\textbf{Shadow estimation.} To estimate cast shadows, we raytrace from each surface point to the light source following the direction of the lighting vectors $\vect{l_m}$ computed above. For each ray we take 16 samples $h$ every 1.5mm starting 3mm away from the start. For all these points, we query the depth of the height map and compute the difference $d_h=z_h-z_s$; if at least one of these differences is negative, there is a shadow. This shadow computation can be differentiably approximated as: $\text{SM}\Big{(} - \text{sigmoid}( d_h)\Big{)}$ (where SM denotes the \textit{softmax} operator).

\subsection{Losses}\label{sec:losses}

Our neural surface is trained with the following losses.
\noindent
\textbf{Angular normal loss.} We apply normal loss on surface normals to match single view normal estimates (from \cite{logothetis2021pxnet}) using the angular loss formula:
\begin{equation}
L_{n}=|\text{atan2}(||\vect{n}_n \times \vect{n}_s||, \vect{n}_n \cdot \vect{n}_s)|\text{max}(\vect{n}_n \cdot \vect{v},0)
\label{eq:angloss}
\end{equation}
For experiments where this loss is used, the relative weighting of this loss is 1 (with normals measured in degrees).

\noindent
\textbf{Rendering loss.} We include an L1 error on the rendered intensities (for all lights $m$) as:  $L_{r}= ||i_{t,m} - i_{r,m}|| $. Relative weights are 100 for LUCES-ST and 1000 for DiLiGenT-MV (image values are rendered in [0,1] and which has DiLiGenT-MV darker images).


\noindent


\begin{table*}[t]
\setlength{\tabcolsep}{3.0pt} 
\begin{center}
\begin{small}  
\begin{tabular}{||c| c c c c c ||c |c|| }
 \hline
 Method & Bear & Buddha & Cow & Pot2 & Reading & Average SE & Median SE \\ 
 \hline\hline

  \hline
  DiLiGenT-MV~\cite{LiZWSDT20} [all views] & ~~~~0.74~~~~ &~~~~0.53~~~~&	~~~~0.83~~~~ &	~~~~0.57~~~~ &	~~~~1.39~~~~ &	~~~~0.81~~~~  & ~~~~0.23~~~~ \\ 
  PS-NeRF~\cite{psnerf} [all views]& 0.45 &  
  0.40 &  0.58 &  0.40 & 0.61 & 0.49  & 0.31 \\ 
\hline
\hline
 Logothetis et al.~\cite{Logothetis22} [1 view - camera 1 only]& 2.62	& 2.77	 & 1.14	 &	0.89	&	6.32	& 2.75 	& 2.41 \\ 
 Logothetis et al.~\cite{Logothetis22} [2 views]& 2.70	& 3.23	& 0.87 &	0.79 &	5.97	&2.71  & 2.51 \\ 

 
PS-NeRF~\cite{psnerf} [2 views] &  2.64 & 1.02 &1.02 & 0.94 &3.88 &	1.90	 &1.57 \\ 


Guo et al.~\cite{guo2022edgepreserving}* [2 views] & 0.86  & 0.51  & 3.21  & 1.39  & 1.05  & 1.40 & 1.18 \\ 

 \hline 
 Ours - [PS-NeRF~\cite{psnerf} normals only]  & 1.17  & 0.57  & 0.82  & 0.78  & 0.90  & 0.85	 &0.63 \\ 


 Ours - [normals only]  & 0.40  & 0.49  & 0.60  & 0.34  & 0.75  & 0.52 	 & 0.38\\ 


Ours - [intensities only] & 0.73  & 0.57  & 0.69  & 0.61  & 0.87  & 0.69  & 0.58\\ 

\hline
~~~~~~Ours  - [PS-NeRF~\cite{psnerf} normals + intensities]~~~~~~ &0.66  & 0.51  & 0.65  & 0.68  & 0.81  & 0.66  & 0.51 \\

Ours  - [normals + intensities ] & 0.57  & 0.51  & 0.75  & 0.56  & 0.75  & 0.63  &0.50 \\ 


 \hline



\end{tabular}
    \caption{This table shows both ablation (last 5 rows) and main results on DiLiGenT-MV~\cite{LiZWSDT20} dataset. For all objects we report the mean shape error as well as average shape error and average median shape error on all objects.  All of our experiments are performed using first 2 views but single view and all view competitors are shown for reference. In the ablation experiment we run our method using single view normal map from  PX-Net~\cite{Logothetis22} (current calibrated PS SOTA) or \cite{psnerf} (originally computed with  \cite{chen2019SDPS_Net} using all 20 views) in order to have a fair comparison with \cite{psnerf}. In addition, the effectiveness of intensity rendering is also ablated for both input normal configurations as well on its own. We note that our best configuration on this experiment is using normals only and significantly outperforms all other 2 view competitors (0.52mm average SE vs 1.4mm for \cite{guo2022edgepreserving}) and its only marginally worse than the 20view SOTA (0.49mm for  \cite{psnerf}). We emphasise that all of our ablation experiments are also significantly outperforming all other competitors showing the strength of our approach. Finally, it is interesting to note that the intensity rendering is only improving performance when combined with \cite{psnerf} normals and it is actually decreasing performance compared to SOTA (PX-Net) normals only. This is probably due to inaccurate near-lighting modeling on DiLiGenT-MV~\cite{LiZWSDT20} as no light angular dissipation factors $\mu$ are provided. In contrast, on the truly near-field LUCES-ST \ref{tab:tab_eval_stereoluces}, intensity rendering improves in most cases.
   }
    \label{tab:Tab_eval_diligent}
\end{small}
\end{center}
\end{table*}

\noindent
\textbf{Depth loss.} Used only at the initialisation stage (as depth estimates are very inaccurate) $z_s$,  $L_{z}=  \lambda _z | z_s - z_t |$. Relative weight is 1 (with depth in mm).


\noindent
\textbf{Regulariser.} For numerical stability reasons, we apply normal and depth regularisers ($\vect{n}=[0,0,1]$, $z=\text{mean}(z_0)$) with respective weights 1e-3 and 1e-4.

\noindent
\textbf{Sample weighting.} To minimise the impact of self-reflections, we note that these tend to occur on oblique points where there are also shadows. Thus, for each point the \textit{approximate ambient occlusion} $a$ (measure of obliqueness, see \cite{Easdon2013AmbientOA}) is computed as $a=\frac{\text{number of shadows}}{\text{number of lights}}$. We multiply normal loss with $a$ and rendering loss by $a^2$ as intensities are less robust to self-reflections than normal estimates 








\noindent
\textbf{Implementation details.} We use official tensorflow implementation of SIREN with $5 \times 512$ layers and 1.05M parameters. For the first layer, we use a 50 frequency in DiLiGenT-MV and 100 in LUCES-ST (due to higher resolution input images).


\subsection{Training}\label{sec:training}

\noindent
\textbf{Initialisation stage.} We note that surface sampling procedure described above requires that the surface is appropriately initialised for shadow computations to be meaningful. To achieve this,  we project the initial surface (obtained from \cite{Logothetis22})  into the unified coordinate system and then pre-train just the SIREN function with normal and depth loss. Of course the initial depth maps are inconsistent and the network at best can converge to an estimate of their average. We also apply data augmentation of $\pm 1mm$ on word coordinate points at that stage in order for the initial surface to be smooth (so that the network does not attempt to make two copies of the surface). We train the pre-initialisation stage for 300 epoch on DiLiGenT-MV and 30 on LUCES-ST which takes around 5 mins on GTX4080 (batchsize 16384). Note that no-ray sampling, rendering and shadows is used in the stage therefore allowing for much higher batchsize and much faster epochs than the main stage. 

\noindent
\textbf{Training stage.} During the main training stage, we use 128 depth samples per ray and 16 shadow samples. We train with batchsize 512 and  1024 on DiLiGenT-MV and LUCES-ST respectively (DiLiGenT-MV has 96 vs 15 lights increasing memory consumption). We run for 300 epochs on DiLiGenT-MV and 50 epochs on LUCES-ST (LUCES-ST objects containing around 1.5M samples with contrast to around 100K for DiLiGenT-MV). To further increase the convergence speed, we only enable rendering and shadows after epoch 2 on LUCES-ST and 10 on DiLiGenT-MV (so the first epochs run with normal loss only and take around half the time). Also see Figure~\ref{fig:roccurve}.

\section{Experimental Setup}
\label{sec:evaluation}
In this section we describe the datasets used in our evaluation, including the new LUCES-ST dataset. We also discuss evaluation metrics and various implementation details.
\subsection{LUCES-ST dataset}
\label{sec:LUCES}
We present LUCES-ST dataset with a subset of 7 objects: \textit{Bell}, \textit{Bunny}, \textit{Cup}, \textit{Hippo}, \textit{Owl}, \textit{Queen}, \textit{Squirrel}, out of the original 14 of~\cite{Logothetis22}. We note that we only reused the objects and their CT scanned GT meshes and all of the stereo capture data is new. We use a stereo capture device with 2x - Flea3 FL3-U3-32S2C-CS 1/2.8" Color USB 3.0 Camera pointgray cameras ($2080\times 1552$~px) and 15 LED lights as shown in Figure~\ref{fig:method}. We use 8mm lenses for the cameras and place the objects around 15-20 cm away from the camera in order for the near lighting effects to be significant (as opposed to DiLiGenT-MV~\cite{LiZWSDT20}). This sparse lighting setting makes the photometric stereo problem extra challenging adding to the value of our dataset. Note that as the 2 stereo images per light are captured simultaneously, the effective light directions for each pixel differ for the 2 views (due to parallax) in contrast to turntable setups like DiLiGenT-MV. This makes application of uncalibrated PS methods (such as \cite{chen2019SDPS_Net}) that rely on the lighting vectors being the same at each view) extra challenging. The data is available to download at \href{https://www.toshiba.eu/pages/eu/Cambridge-Research-Laboratory/luces}{https://www.toshiba.eu/pages/eu/Cambridge-Research-Laboratory/luces}.

\begin{figure*}[t]
    \begin{center}\includegraphics[width=0.85\textwidth]{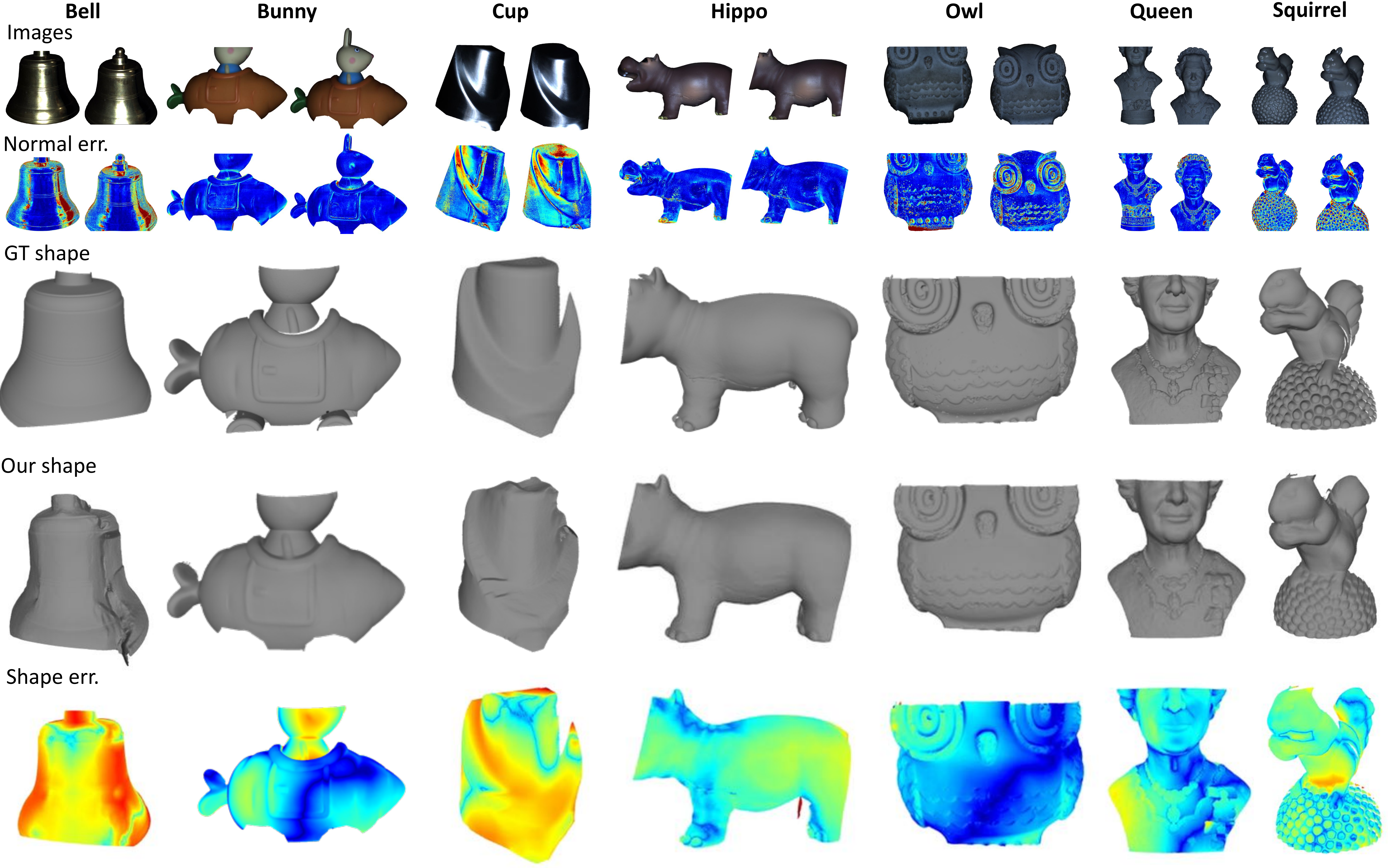}\end{center}
    \vspace{-0.6cm}
    \caption{This figure shows the qualitative results of our method on LUCES-ST dataset. The first thre rows show the cropped images and corresponding error images of normals estimated from PX-Net~\cite{Logothetis22} and the ground truth shape. The final two rows show the shape predicted by our method and corresponding error map (from ground truth to reconstruction). As in Figure~\ref{fig:stereoresults} any errors larger than 1.5mm are clamped to a dark red color. Our method performs well on \textit{Hippo}, \textit{Owl}, \textit{Queen} and \textit{Squirrel} objects as also reflected in Table~\ref{tab:tab_eval_stereoluces}. Performance is worse on highly specular, metallic objects such as \textit{Bell} and \textit{Cup}.}
    \label{fig:stereoresultsluces}
\end{figure*}

\subsection{Metrics and comparison details}

\noindent
\textbf{Performance metrics.} As we only use a stereo pair of views to compute reconstructions, full object Hausdorff distances are not informative. Therefore, in order to have a fair comparison, we compute a cropped ground truth (though back-projection of the ground truth depth maps) and the compute Hausdorff distance \textit{from the ground truth} to the reconstructed objects. That makes sure that the metric is fair for all competitors producing variable size outputs. Therefore, computed error maps are all shown on the GT and hence are comparable between different competitors.

\noindent
\textbf{Datasets.} Along with LUCES-ST, we evaluate our approach on the synthetic version of LUCES-ST (see Table~\ref{tab:tab_eval_stereoluces} and supplementary material) as well as a popular real multi-view PS benchmark - DiLiGenT-MV~\cite{LiZWSDT20}. DiLiGeNT contains 5 objects \textit{Bear}, \textit{Buddha}, \textit{Cow}, \textit{Pot2} and \textit{Reading} captured from 20 views, with 96 light images of $612 \times 512$ resolution. The objects are around 1.5~m away from all cameras (turntable capture setup) and the camera focal length of 50~mm approximates orthographic viewing. As we are only interested in a single pair of views, we chose the first 2 views.

\noindent
\textbf{Adapting competing methods to binocular PS setup.} No recent method has focused on the Binocular PS problem so fair comparison is non trivial. We compare with \cite{Logothetis22} which shows SOTA performance on single view and to compute a 2 view result, the 2 independent view reconstructions are concatenated and merged with Poisson reconstruction \cite{kazhdan2006poisson}. The very recent single view method of \cite{guo2022edgepreserving} is conceptually our closest match due to the use of the SIREN surface and rendering. To simulate their 2 view extension, we adapt our implementation to perform pure Lambertian rendering and disabled the normal loss and shadow computation. We note that the same initialisation with normals and average surface that we used is also used for them. Finally, we compare to PS-NeRF~\cite{psnerf} which has available code online. We note that the surface is not updated during their second stage of training (which optimises re-rendering) so for the purpose of surface error, stage 1 is only used. That means that normal maps and average intensity images are only used. Therefore, to apply them on LUCES-ST, the normal maps of \cite{Logothetis22}  are used.

\section{Experiments}
\label{sec:experiments}

In this section we report results on DiLiGenT-MV~\cite{LiZWSDT20} and LUCES-ST datasets. This is shown quantitatively on Tables~\ref{tab:Tab_eval_diligent} and \ref{tab:tab_eval_stereoluces} and visualised in Figures \ref{fig:stereoresults} and \ref{fig:stereoresultsluces} respectively. Also the use of intensity rendering is ablated for both datasets by provided results with normal only, intensity only and combined losses. 

\begin{table*}[ht]
\setlength{\tabcolsep}{3.0pt} 
\begin{center}
\begin{small} 
\begin{tabular}{|c | c| c c c c c c c | c c| } 
 \hline
Method & Synthetic / Real & Bell  & Bunny & Cup & Hippo  & Owl  & Queen  & Squirrel  & Average SE & Median SE \\ 
\hline
Ours - [normals only] & Synthetic & 0.90  & 1.37  & 1.18  & 0.79  & 0.78  & 0.46  & 0.60  & 0.87  & 0.57 \\ 

Ours - [intensities only]& Synthetic &  0.73  & 0.70  & 1.24  & 0.36  & 0.36  & 0.32  & 0.27  & 0.57  & 0.24 \\


\hline
Ours - [normals + intensities ]& Synthetic &0.75  & 0.82  & 1.20  & 0.76  & 0.70  & 0.44  & 0.55  & 0.75 & 0.51 \\

\hline
\hline
 Logothetis et al.~\cite{Logothetis22}  [2 views] & Real & ~~1.24~~	 &	~~1.14~~	&  ~~0.69~~	&	~~0.62~~	 & ~~0.64~~	&~~0.97~~	& ~~1.92~~	& ~~1.03~~  &  ~~0.78~~\\
PS-NeRF~\cite{psnerf}& Real & 1.35 &	0.76	& 1.11	&1.40	 &0.88	 &0.58	& 0.95	&1.00 &  0.85\\ 
Guo et al.~\cite{guo2022edgepreserving}* & Real & 4.53  & 1.10  & 3.41  & 0.71  & 0.38  & 0.61  & 0.85  & 1.66 & 1.83 \\
\hline
Ours - [GT normals only]& Real & 0.18  & 0.71  & 0.34  & 0.23  & 0.18  & 0.28  & 0.41  & 0.33  & 0.25 \\ 
Ours - [normals only]& Real & 1.5  & 0.92  & 1.65  & 1.10  & 0.49  & 0.62  & 0.61 & 0.98   & 0.77 \\ 
%
Ours - [intensities only]& Real &1.87  & 1.11  & 1.33  & 0.50  & 0.35  & 0.63  & 0.55  & 0.91 & 0.80 \\
\hline

Ours - [normals + intensities]& Real & 1.41  & 1.01  & 1.74  & 1.05  & 0.46  & 0.62  & 0.59  & 0.98 & 0.77 \\

 \hline
\end{tabular}
\end{small} 
\end{center}
   \vspace{-0.4cm}
    \caption{This figure shows the quantitative results of our method  on LUCES-Stereo dataset. We show comparisons with \cite{Logothetis22},\cite{psnerf} and \cite{guo2022edgepreserving} (adapted to Binocular Photometric Stereo setup) as well as ablations of the use of intensity rendering. In addition, we also provide an ablation of our method on synthetic version of  LUCES-Stereo dataset (containing Blender renderings of the same objects with different pose and segmentation masks, see supplementary Figure 1). Finally, results using ground truth normals as an input are also shown to provide an estimate of the best achievable error of our method. It is noted that the best overall configuration in term of mean error is using intensities loss only with normal loss improving performance in some objects and decreasing in some others.    
  }
    \label{tab:tab_eval_stereoluces}
\end{table*}

\noindent
\textbf{DiLiGenT-MV~\cite{LiZWSDT20} experiments.} We note that our best configuration on DiLiGenT-MV~\cite{LiZWSDT20} (see Table~\ref{tab:Tab_eval_diligent}) turns out to be using normal loss only and significantly outperforms all other 2 view competitors (0.52mm vs 1.4mm average SE for \cite{guo2022edgepreserving}) and is only marginally worse than~\cite{psnerf} using 20 views (0.49mm). In addition, we note that all of our configuration experiments with both sets of normal map inputs (ie. normals used by PS-NeRF~\cite{psnerf} vs PX-Net~\cite{Logothetis22}) and with/without intensity also outperform all other competitors and achieve less that 1mm in almost all experiments.  It is also notable that the use of intensity rendering seems to degrade the performance when combined with  PX-Net normals but offers a small improvement when combined with the less accurate \cite{psnerf} normal input. 
In contrast, in the LUCES-ST experiments (see bellow) intensity rendering offers a clear advantage. 

The degradation of performance using intensity rendering on  DiLiGenT-MV~\cite{LiZWSDT20} can be attributed to potentially inaccurate near-lighting modeling, as no light angular dissipation factors $\mu$ are provided. In fact, DiLiGenT-MV~\cite{LiZWSDT20} provides point light positions and far-field equivalent light intensities, therefore to apply the near-field lighting model ,$\mu =0$ was assumed and light intensities were compensated with inverse square of average object distance. The last step can be inaccurate depending on exactly how the far-field equivalent light intensities were measured. In contrast, normal estimation networks can be robust to miss-calibration though data augmentation (e.g. PX-Net~\cite{Logothetis22}) or be outright self-calibrating (e.g. \cite{chen2019SDPS_Net}) and do not suffer from the aforementioned issue.
\noindent
\textbf{LUCES-ST experiments.} LUCES-ST experiments are shown quantitatively in Table \ref{tab:tab_eval_stereoluces} and qualitatively  in Figure \ref{fig:stereoresults}. Similar to DiLiGenT-MV~\cite{LiZWSDT20} experiments, the use of intensity rendering is also ablated an in fact it does seem to have an increase of performance for most experiments. 

The best configuration turns out to be \textit{intensity only} in terms of mean error (0.91mm) with \textit{normals + intensity} and \textit{normals only}  being slightly better in terms of median error (0.77mm vs 0.8mm). PS-NeRF~\cite{psnerf} and  Logothetis et al.~\cite{Logothetis22} are slightly worse with mean errors of (1.0mm and 1.03mm) respectively.  We note that all methods in this dataset are using the single -view normal predictions from \cite{Logothetis22} (even \cite{guo2022edgepreserving} was ran with 2 epochs of normal loss only for initialisation) therefore the spread of results is much less that in DiLiGenT-MV. In addition, since accurate, near-field light calibration is available,  intensity rendering  improves performance on most experiments. A notable exception is the metallic \textit{Bell} which contains environment reflections which are not modelled in the assumed rendering process.  

To further re-enforce the usefulness of intensity rendering, we also add Blender renderings of the same objects with a reasonable guess of their materials (metallic \textit{Bell}, ceramic \textit{Owl}, porcelain \textit{Squirrel}, etc). Note that the poses and segmentation masks are different therefore synthetic to real comparison is not fair. Nevertheless, intensity offers a clear advantage for all objects and performs the best on average with a significant margin (0.57mm intensity only vs 0.87mm normals only vs 0.75mm combined). In the real experiments, the superiority of intensity rendering is not always true as real data may contain totally un-modelled effects such as ambient light (\cite{logothetis2016near}), camera noise, and even a small calibration error. Normal estimation networks are very robust to a lot of effects and thus offer a useful source of information for real world experiments. 

An additional considerations is that DiLiGenT-MV~\cite{LiZWSDT20} contains 96 lights whereas LUCES-Stereo only uses 15. This gives a significant advantage to normal estimation networks that can use all of the lights to gain robustness to real world imperfections. In contrast, averaging the rendering loss (even L1) over a set of lights is more susceptible to clear outliers (and in fact the more lights, the higher the chance that one of the lights contains un-modelled effects such as self reflections).

Finally, to calibrate the limit of precision of out method, results with ground truth normals are also included for LUCES-Stereo (line 7 in Table~\ref{tab:tab_eval_stereoluces}). The obtained error of 0.33mm is much smaller than any real experiment (0.91mm) but certainly non-negligible, potentially signifying the need for higher learning capacity network.

We note that some additional visualisations including re-renderings and recovered BRDFs LUCES-Stereo experiments are available in the supplementary.

\section{Conclusion}
\label{sec:conclusion}

In this work we propose a novel neural heightmap approach to Binocular Photometric Stereo along with a new dataset - LUCES-Stereo. We show that our approach is able to extract accurate shape from extremely sparse views (i.e. 2 views) significantly better than single view photometric stereo~\cite{Logothetis22} and even reach similar performance in terms of average shape error to the state-of-the-art multi-view photometric stereo method~\cite{psnerf} on DiLiGeNT~\cite{LiZWSDT20} benchmark.   


{\small
\bibliographystyle{ieee_fullname}
\bibliography{egbib}
}

\appendix

\section{Appendix}

 This appendix presents supplementary material for our main submission. In Section ~\ref{sec:luces} we provide additional details for our newly introduced Binocular Photometric Stereo LUCES-ST dataset.
Section \ref{sec:brdf} provides some additional details about the learned BRDF renderer.

\begin{table*}
    \begin{center}
        \begin{tabular}{|c | c c c c c c c | c |} 
 \hline
Object &  Bell  & Bunny  & Cup & Hippo  & Owl  & Queen  & Squirrel  & Average \\
\hline
View 1 angular error (degrees) & 15.12 &	6.08	 &13.80	 &5.88 &	10.75	 &9.65 &	13.76	 & 10.21 \\
View 2 angular error (degrees) & 15.16 &	5.71	 &15.98 &	7.07	& 11.94 &	9.21 &	13.60 &	10.85 \\
\hline
\end{tabular}
  \end{center}
    \caption{Evaluation of the accuracy of the normals predicted by~\cite{Logothetis22} on images of LUCES-stereo dataset. Not surprisingly, this is not negligible as there are only 15 lights available and the object's geometry is challenging  (causing shadows, self reflections etc). The normal error maps are shown in Figure 5 of the main paper.}
    \label{tab:tab_eval_normals_stereoluces}  
\end{table*}

\begin{figure*}[!ht]
    \includegraphics[width=\textwidth]{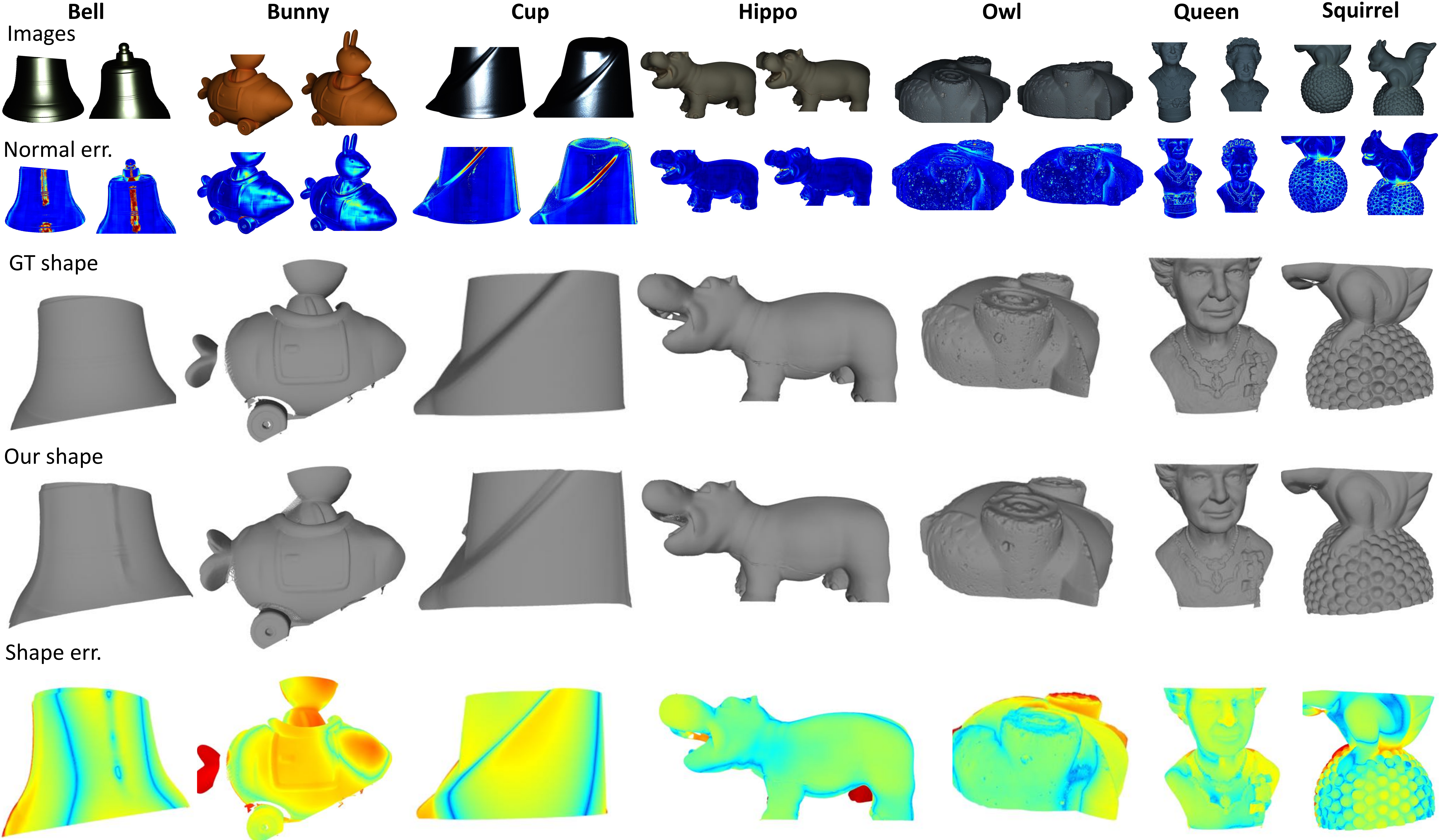}
    \caption{This figure shows the qualitative results of our method (normals+intenisty loss variation) on LUCES-ST-synthetic dataset. This is a synthetic counterpart to Figure 5 of the main text. The first three rows show the cropped images and corresponding error images of normals estimated from PX-Net~\cite{Logothetis22} and the ground truth shape. The final two rows show the shape predicted by our method and corresponding error map (from ground truth to reconstruction). Similarly to Figures 4 and 5 of the main text, any errors larger than 1.5mm are clamped to a dark red color. In contrast to real data results, out method performs relatively well on all objects (performance drop on  metallic objects such as Bell and Cup is less significant than in the real data) with the challenging geometry regions (around occlusion boundaries) accumulating most of the error.}    
    \label{fig:blender_results}
\end{figure*}

\section{LUCES-Stereo}
\label{sec:luces}

\begin{figure*}[!ht]
    \includegraphics[width=\textwidth]{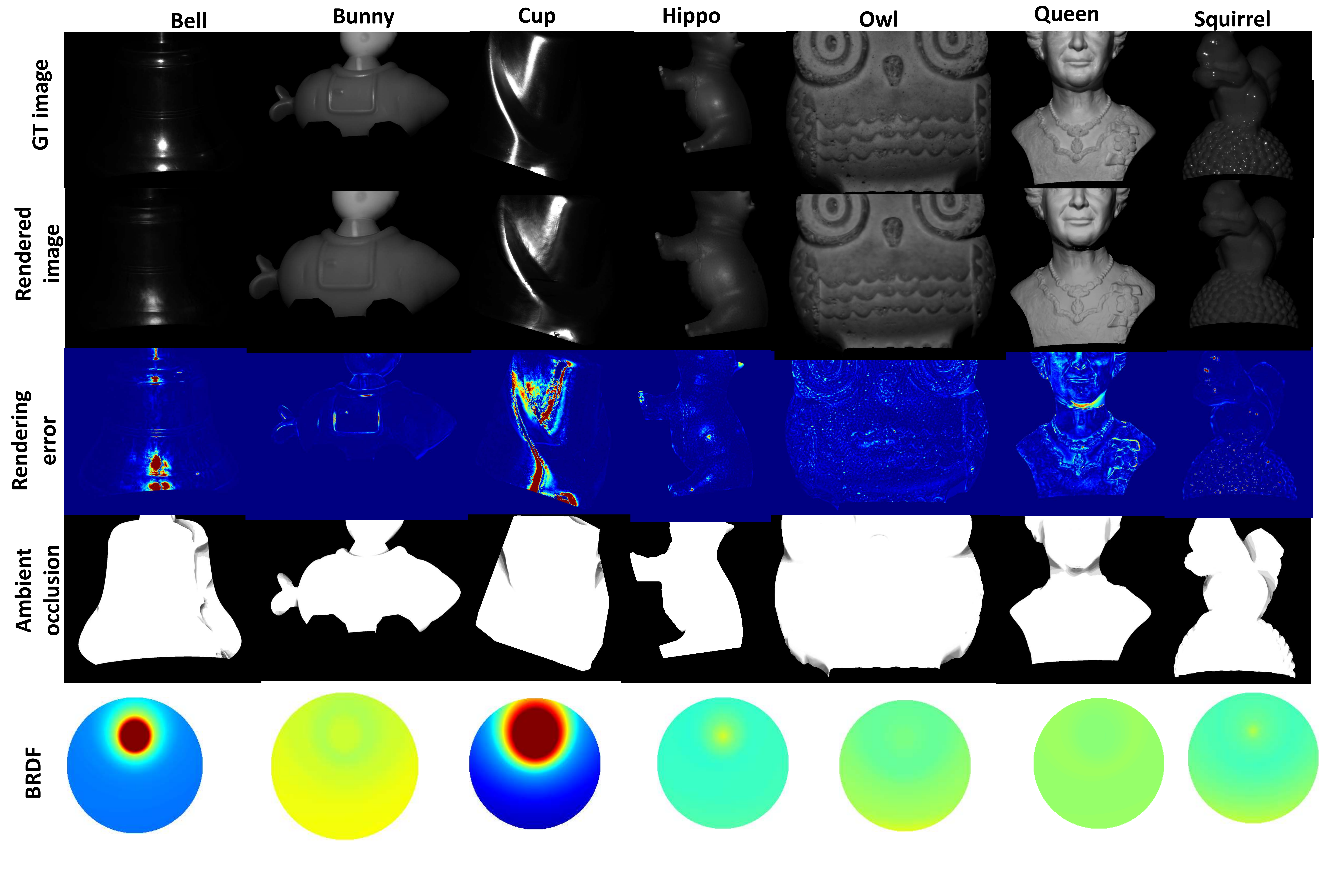}
    \caption{Rendering visualisation for all LUCES-ST objects. From top to bottom, rows include ground truth first image of first camera, respective rendered image, rendering error (with dark red corresponding to 0.25), ambient occlusion (number of shadows/number of images) and recovered BRDF. We note that  the square of the ambient occlusion is used to weigh the rendering loss, hence rendering error in these regions does not really affect the training procedure. We also note that the BRDF is visualised excluding the $(\vect{n}\cdot \vect{l})$ component (hence the values around the edges of the sphere are meaningless). Moreover, this BRDF rendering corresponds to $\vect{v}=[0,0,1]$ (i.e. `orthographic viewing') and  $\vect{l}=[0,\sqrt{2}/2,\sqrt{2}/2]$ and the colormap is chosen with green corresponding the $1$ and dark red to $>2$.  }
    \label{fig:stereoresultsluces_rend}
\end{figure*}

First, we note that we only reused 7 of the original LUCES~\cite{Logothetis22} objects and their CT scanned GT meshes and all of the stereo capture data are new. We chose a variety of materials: \textit{Bell} is bronze, \textit{Bunny} is shiny plastic,  \textit{Cup} is aluminium, \textit{Hippo} is plastic, \textit{Owl} is ceramic, \textit{Queen} is plaster and \textit{Squirrel} is porcelain. In fact, the stereo capture device only contains 15 lights as the data acquisition speed is very important in an industrial inspection setting.  This sparse lighting setting makes the photometric stereo problem highly challenging adding to the value of our dataset. Note that as the 2 stereo images per light are captured simultaneously, the effective light directions for each pixel differ for the 2 views (due to parallax) in contrast to turntable setups like DiLiGenT-MV. This makes the application of uncalibrated PS methods (such as \cite{chen2019SDPS_Net}) that rely on the lighting vectors being the same at each less practicable. Indeed, calibrated PS SOTA normal estimation on these objects (using \cite{Logothetis22}, see Table~\ref{tab:tab_eval_normals_stereoluces}) achieves a non negligible error (i.e. mean $10.5^o$). We hope that our PS data will be useful for future research on sparse PS aiming to minimise this error.  


Similarly to competing datasets (original LUCES~\cite{Logothetis22}, DiLiGenT~\cite{LiZWSDT20})  the camera intrinsics and baseline were computed using a the standard checkerboard calibration target procedure and the lighting calibration parameters (using the near lighting model of \cite{Mecca2014near} that includes position, brightness, principal direction and attenuation factor as explained in Section 3 of the main paper) using a diffuse calibration target\footnote{\href{https://www.edmundoptics.co.uk/f/white-balance-reflectance-targets/13169/}{https://www.edmundoptics.co.uk/f/white-balance-reflectance-targets/13169/}}. We note that the target was captured in 5 distances of 22, 24, 25, 29, 30 cm, thus providing a total of 5x2x15=150 calibration images. The lighting model parameters were fitted with differentiable rendering obtaining a final re-rendering error of $\approx0.005$, thus the expected accuracy of the light calibration should be around 0.5\%. We note that despite the fact that the images of the stereo pair are captured simultaneously, the effective brightness of the LEDs are not identical to both views due to different camera sensitivity (which is also channel dependent). This makes uncalibrated PS especially challenging as both light positions and orientations (from the cameras point of view) and brightness different between respective stereo pair cameras.

The ground truth meshes are also aligned with Meshlab\footnote{\href{ https://www.meshlab.net/}{ https://www.meshlab.net/}} and thus ground truth normals and depth are rendered (with Blender\footnote{\href{https:https://www.blender.org/}{https:https://www.blender.org/}}) for each view. In addition, we note that segmentation masks were manually edited to exclude some points that the 'GT' may be unreliable such as the wheels at the bottom of Bunny. In addition, for the making a fair evaluation of stereo methods, the segmentation masks were further cropped to only include points on the field of views of both cameras (but not necessarily co-visible due to self occlusions). 

\noindent
\textbf{Synthetic LUCES-ST.} To further investigate synthetic to real gap, we also rendered all 7 objects with Blender using the same 15 stereo lights setup at the real one and with reasonable material guesses. This is shown in Figure \ref{fig:blender_results} along with the results of the normals + rendering variation.  Note however that the poses are not identical to the real data.

\section{BRDF Renderer}
\label{sec:brdf}

This section provides additional details about the learnt BRDF renderer. First of all, we note that considering RGB images has very little value compared to grayscale ones (despite the added computation and memory overhead), especially since RGB cameras usually have Bayer pattern filters and the RGB color are recovered using demosaicing, we which we optimised to mostly preserve brigthness and not `true' color. Therefore, we follow standard RGB to gray conversion\footnote{\href{https://docs.opencv.org/3.1.0/de/d25/imgproc_color_conversions.html}{https://docs.opencv.org/3.1.0/de/d25/imgproc\_color\_conversions.html}} on input images, and optimise our RGB renderer using scalar albedos and intenisty rendering.

\noindent
\textbf{BRDF parameterisation.}
Our aim is to learn a single BRDF model (assuming uniform material for the whole object), following the principles describe in the MERL real material database \cite{Matusik2003jul}. First, we note that in the case of normal, viewing and lighting vectors having relative angles $>90^o$, the reflected light is always 0, therefore any real rendering should include a $\text{sign} (\vect{n}\cdot \vect{l}) \text{sign} (\vect{n}\cdot \vect{v}) \text{sign} (\vect{v}\cdot \vect{l}) $ (sign is a binary flag 0 for negatives, 1 for positives and was ommited from equation in Line 396 of main text for clarity).
In addition, we note that the BRDF is only a function of the relative angles of these vectors, so they can all be rotated such as $\vect{n}=[0,0,1]$ (by rotating around the $\vect{n} \times [0,0,1]$ axis with angle $\arccos (\vect{n} \cdot [0,0,1])$). Thus $\text{BRDF} (\vect{n},\vect{l},\vect{v}) \vcentcolon = \text{BRDF} (\vect{l_n},\vect{v_n})$ with $\vect{l_n},\vect{v_n}$ the new rotated vectors. In addition, as most of the specular lobe is around the half vector  $\vect{h}=\frac{\vect{l_n}+\vect{v_n}} {|\vect{l_n}+\vect{v_n}|} $, \cite{Matusik2003jul} recommends parameterising $\vect{h}$, as well as the difference vector $\vect{d}$ between  $\vect{h}$ and  $\vect{l_n}$. In fact,  $\vect{d}$ is computed by rotating both $\vect{h}$ and $\vect{l_n}$ such as  $\vect{h}$ is aligned with $[0,0,1]$. Thus the 4 angles defining the BRDF can now be computed as (using superscripts to denote $x,y,z$ components of 3D vectors):

\begin{itemize}
    \item $\theta _h= \arccos (h^z) \in [0,\pi/2]$.
    \item $\phi _h= \text{atan2}(h^y, h^x) \in [-\pi,\pi]$.
     \item $\theta _d= \arccos (d^z) \in [0,\pi/2]$
    \item $\phi _d= \text{atan2}(d^y,d^x) \in [-\pi,\pi]$
\end{itemize}

In addition, any real BRDF must follow the Helmholtz reciprocity constraint which enforces symmetry between $\vect{l_n}$ and $\vect{v_n}$ vectors; in the 4 angle parameterisation that translates to periodicty wrt to $\phi _d$ with period $\pi$ (instead of $2\pi)$, therefore from a learning perspective $\phi _d  \in [0,\pi]$ is sufficient. Moreover, since we aim to learn these kind of BRDFs from limited image data, it is preferred to only only consider isotropic materials (such as the 100 ones in  \cite{Matusik2003jul}) to minimise the overfitting chances. Thus, this corresponds to ignoring $\phi _h$. Finally, to simplify the learning procedure, we explicitly factor out the the incident light component  $(\vect{n}\cdot \vect{l})$ and thus learn  $\text{BRDF}  \vcentcolon =(\vect{n}\cdot \vect{l_m}) \text{MLP} (\theta _h, \theta _d, \phi _d)$ (in fact depending on the literature sources, the BRDF is usually defined as the remaining component after  $(\vect{n}\cdot \vect{l})$ is factored out).

For the MLP component, we use three fully connected layers containing 16 units each with relu activation followed by one fully connected with a single value output and exponential activation function. That forces the values to be always non-zero and encourages them to be  around 1 which corresponds to Lambertian reflectace\footnote{Advanced graphics models computing multiple light bounches include a $1/\pi$ factor for energy conservation but that is beyond the scope of out work since we do not compute self reflections and thus any scaling factor is meangless.} and it is a reasonable mean value.

Visualisation of the renderings and recovered BRDFs for all LUCES-ST objects are shown in Figure~\ref{fig:stereoresultsluces_rend}. It is observed that for the metallic objects (Bell, Cup) very narrow specular lobes are recovered; specular dielectric materials (Bunny, Hippo and Squirrel) have less peaky specular lobes and for mostly diffuse objects (Owl,Queen) a mostly diffuse BRDF is recovered.

\end{document}